\pdfoutput=1
\documentclass[conference]{IEEEtran}
\IEEEoverridecommandlockouts
\usepackage{cite}
\usepackage{amsmath,amssymb,amsfonts}
\usepackage{algorithmic}
\usepackage{graphicx} 
\usepackage{textcomp}
\usepackage{xcolor}
\usepackage{microtype}
\usepackage{subcaption} 
\usepackage{rotating} 
\usepackage{array}
\usepackage[T1]{fontenc}
\usepackage{booktabs}  
\usepackage{multirow}  
\usepackage[export]{adjustbox} 

\usepackage[colorlinks, linkcolor=black, anchorcolor=black, citecolor=blue]{hyperref} 
\def\BibTeX{{\rm B\kern-.05em{\sc i\kern-.025em b}\kern-.08em
    T\kern-.1667em\lower.7ex\hbox{E}\kern-.125emX}}

\begin{document}

\title{Decoupling Numerical and Structural Parameters: An Empirical Study on Adaptive Genetic Algorithms via Deep Reinforcement Learning for the Large-Scale TSP}
\author{\IEEEauthorblockN{Hongyu Wang}
\IEEEauthorblockA{\textit{School of Mathematics and Statistics} \\
\textit{Xi'an Jiaotong University}\\
Xi'an, China \\
3121670564@stu.xjtu.edu.cn}
\and
\IEEEauthorblockN{Yuhan Jing}
\IEEEauthorblockA{\textit{School of Mechanical Engineering} \\
\textit{Xi'an Jiaotong University}\\
Xi'an, China \\
frankjing420@stu.xjtu.edu.cn}
\and
\IEEEauthorblockN{Yibing Shi}
\IEEEauthorblockA{\textit{School of Mathematics and Statistics} \\
\textit{Xi'an Jiaotong University}\\
Xi'an, China \\
1612l@stu.xjtu.edu.cn}
\and
\IEEEauthorblockN{Enjin Zhou}
\IEEEauthorblockA{\textit{School of Mathematics and Statistics} \\
\textit{Xi'an Jiaotong University}\\
Xi'an, China \\
GINzej@stu.xjtu.edu.cn}
\and
\IEEEauthorblockN{Haotian Zhang}
\IEEEauthorblockA{\textit{Frontier Institute of Science and Technology} \\
\textit{Xi'an Jiaotong University}\\
Xi'an, China \\
ht.zhang@xjtu.edu.cn}
\and
\IEEEauthorblockN{Jialong Shi\thanks{Corresponding author: Jialong Shi (jialong.shi@xjtu.edu.cn). This work was supported in part by the National Natural Science Foundation of China (Grants 12571590 and 12401670), the fellowship of China National Postdoctoral Program for Innovative Talents (Grant BX20240284), and the fellowship from the China Postdoctoral Science Foundation (Grant 2025M771731).}}
\IEEEauthorblockA{\textit{School of Mathematics and Statistics} \\
\textit{Xi'an Jiaotong University}\\
Xi'an, China \\
jialong.shi@xjtu.edu.cn}
}
\maketitle

\begin{abstract}
Proper parameter configuration is a prerequisite for the success of Evolutionary Algorithms (EAs). While various adaptive strategies have been proposed, it remains an open question whether all control dimensions contribute equally to algorithmic scalability. To investigate this, we categorize control variables into \textit{numerical parameters} (e.g., crossover and mutation rates) and \textit{structural parameters} (e.g., population size and operator switching), hypothesizing that they play distinct roles. This paper presents an empirical study utilizing a dual-level Deep Reinforcement Learning (DRL) framework to decouple and analyze the impact of these two dimensions on the Traveling Salesman Problem (TSP). We employ a Recurrent PPO agent to dynamically regulate these parameters, treating the DRL model as a probe to reveal evolutionary dynamics. Experimental results confirm the effectiveness of this approach: the learned policies outperform static baselines, reducing the optimality gap by approximately 45\% on the largest tested instance (rl5915). Building on this validated framework, our ablation analysis reveals a fundamental insight: while numerical tuning offers local refinement, \textit{structural plasticity} is the decisive factor in preventing stagnation and facilitating escape from local optima. These findings suggest that future automated algorithm design should prioritize dynamic structural reconfiguration over fine-grained probability adjustment. To facilitate reproducibility, the source code is available at \url{https://github.com/StarDream1314/DRLGA-TSP}
\end{abstract}

\begin{IEEEkeywords}
Hyper-heuristics, Genetic Algorithms, Deep Reinforcement Learning, Traveling Salesman Problem, Adaptive Parameter Control, Zero-Shot Generalization.
\end{IEEEkeywords}

\section{Introduction}

Hyper-heuristics and Evolutionary Algorithms (EAs) are robust methodologies for solving complex NP-hard combinatorial optimization problems. However, their performance remains notoriously sensitive to parameter configurations \cite{burke2013hyper, huang2020survey}. Traditionally, Adaptive Parameter Control (APC) relied on manually designed rules or fuzzy logic to adjust parameters online \cite{Farinati2024DynamicPopSurvey, Subburaj2025FuzzyMemetic}. As problem scales have grown, the field has shifted towards Automated Dynamic Algorithm Configuration (DAC), modeling parameter control as a sequential decision-making process \cite{adriaensen2022automated, Talbi2021MLMetaheuristics}. Deep Reinforcement Learning (DRL) has emerged as a leading approach in this domain, leveraging its ability to map high-dimensional optimization states to adaptive control actions \cite{shala2020learning, berto2023rl4co}.

Despite DRL's promise, a critical gap persists in the literature. Most existing RL-for-EA frameworks focus almost exclusively on fine-tuning \textit{numerical} parameters (e.g., crossover and mutation rates) within a static algorithmic scaffold \cite{woodcock2025gaq, zhou2024moea}. Alternatively, Neural Combinatorial Optimization (NCO) approaches use RL merely to initialize solutions before EA refinement, rather than controlling internal evolutionary dynamics \cite{iet2024tsp, bello2016nco, li2025bridging_ea_rl}. These methods largely overlook \textit{structural plasticity}—the dynamic adjustment of fundamental algorithmic components like population sizing and operator topologies. Theoretical analyses show that rigid structural configurations are fragile; slight deviations in problem characteristics can degrade algorithmic performance exponentially \cite{Fragility2026, NetworkInsights2025}. A solver optimized for one instance often fails catastrophically on others because fine-tuning probabilities cannot overcome the bottleneck of a rigid population structure.

To address this rigidity, we propose a Dual-Level Deep Reinforcement Learning Genetic Algorithm (DRLGA). By treating structural parameters as first-class, learnable control variables alongside numerical ones, we formulate a unified closed-loop controller. We model the optimization as a Partially Observable Markov Decision Process (POMDP) and employ a \textbf{Recurrent Proximal Policy Optimization (Recurrent PPO)} agent equipped with Long Short-Term Memory (LSTM) to capture non-Markovian evolutionary histories. The agent acts as a high-level scheduler, regulating both numerical probabilities and structural modes, which are then executed by a high-performance evolutionary worker.

\begin{figure}[htbp]
    \centering
    \includegraphics[width=\linewidth]{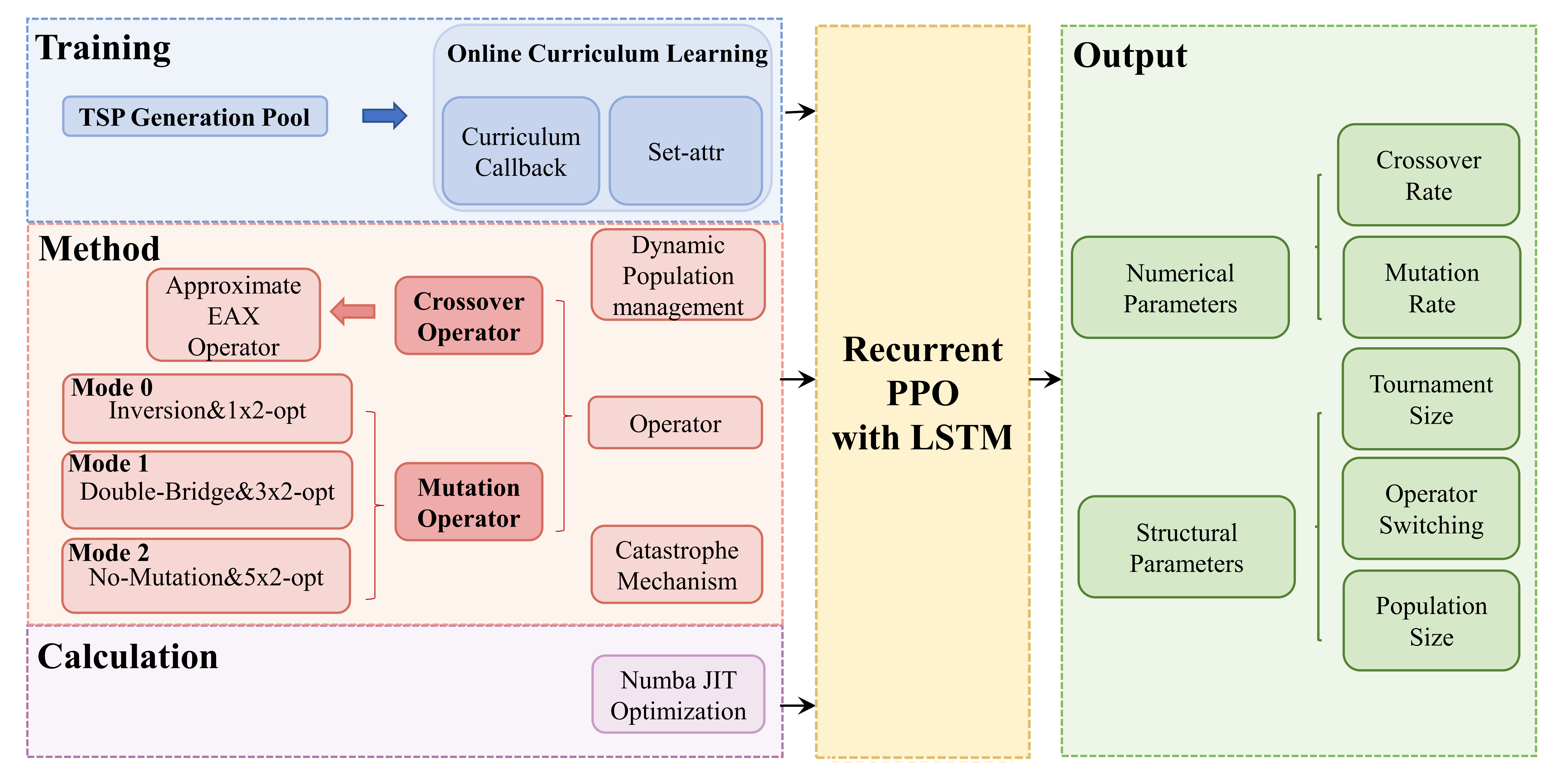}
    \caption{A Dual-level Adaptive Control Architecture}
    \label{fig:dual_level_architecture}
\end{figure}

The overall framework of our proposed model is illustrated in Figure \ref{fig:dual_level_architecture}. 

Our contributions are summarized as follows:
\begin{enumerate}
    \item \textbf{Dual-Level Control Framework:} We propose DRLGA, which synergizes dynamic population management and operator switching (structural) with traditional probability adjustments (numerical), significantly enhancing search versatility across varied fitness landscapes.
    \item \textbf{Robust Zero-Shot Generalization:} Leveraging an LSTM-based Recurrent PPO scheduler trained solely on small-scale synthetic data (up to 1200 cities), our agent successfully generalizes zero-shot to massive benchmarks. It reduces the optimality gap by approximately 45\% on the 5915-city instance (rl5915), achieving sustained stepwise descent where static methods stagnate.
    \item \textbf{Decoupling Structural vs. Numerical Impact:} Through systematic ablation, we reveal that while numerical tuning aids local convergence, structural plasticity is the decisive factor in preventing stagnation in high-dimensional spaces, offering a new paradigm for automated algorithm design.
\end{enumerate}

\section{Methodology Overview}
This paper formulates the adaptive control problem within Genetic Algorithms as a Partially Observable Markov Decision Process (POMDP). In this framework, the state $S_t$ consists of a set of features describing the current runtime status of the GA. This includes 4 static features (problem size, aspect ratio, X distribution, Y distribution) and 11 dynamic features (time progress, optimality gap, recent improvement, stagnation counter, population average gap, population variance, last action: CR, last action: MR, last action: Op Mode, current population size, coefficient of variation); the action $A_t$ is a composite vector containing both numerical parameters (crossover rate, mutation rate) and structural parameters (population size, operator mode, tournament size); the optimization objective is to maximize the long-term improvement of solution quality within a given generation, which translates to minimizing the TSP path length.

\begin{figure}[htbp]
    \centering
    \includegraphics[width=0.6\linewidth]{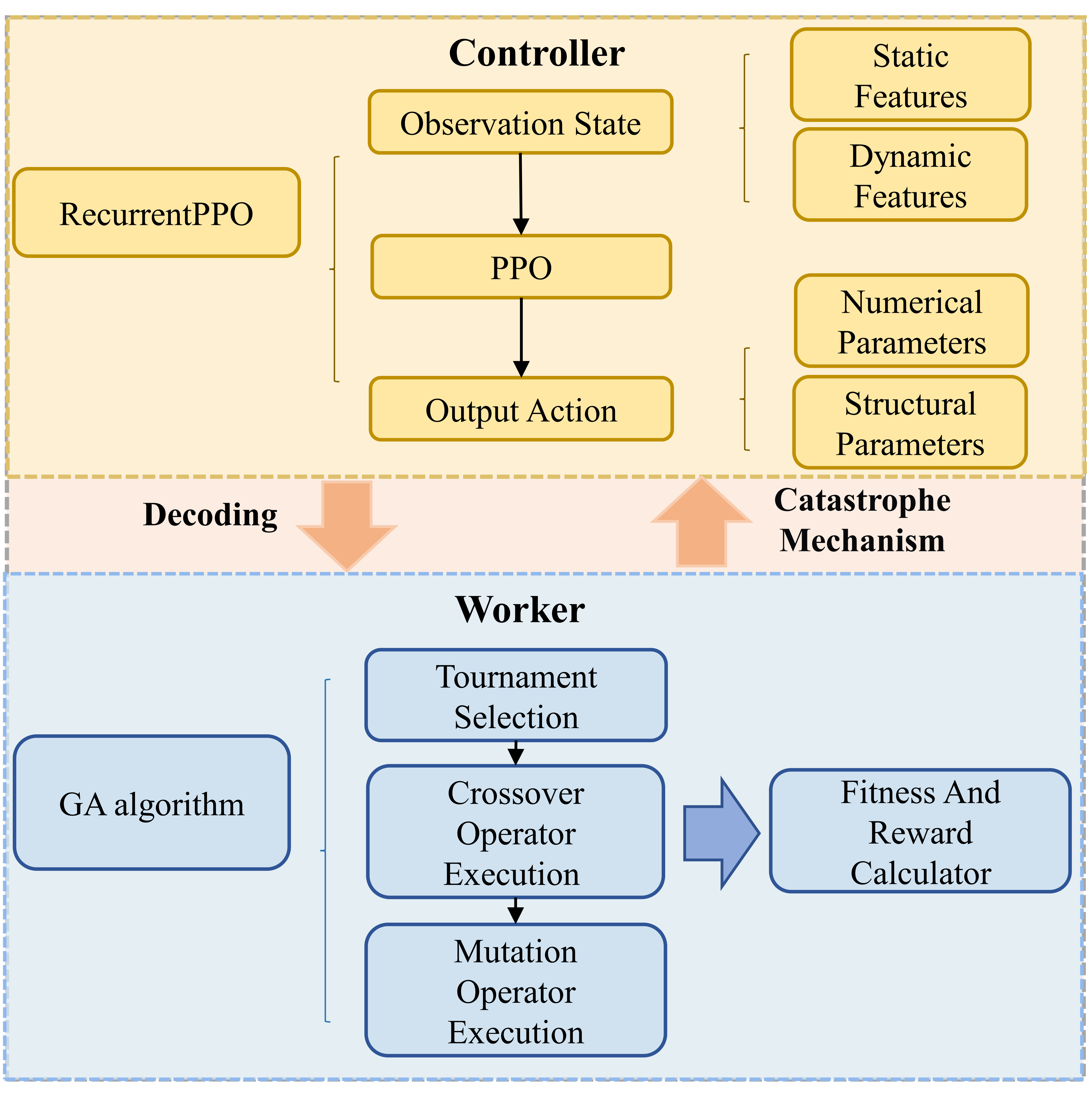}
    \caption{The Upper Controller and The Lower Worker}
    \label{fig:controller_worker_architecture} 
\end{figure}

Building upon these definitions, we construct a dual-level adaptive control architecture. The upper layer is the "Controller", where a Deep Reinforcement Learning agent observes the evolutionary state and outputs high-level control strategies. The lower layer is the "Worker", where a high-performance evolutionary algorithm executes specific search tasks using the specified parameters and feeds back the execution results. Information flows cyclically between the upper and lower layers: the Controller generates action $A_t$ based on the current state $S_t$, and the Worker executes $A_t$, causing the state to evolve to $S_{t+1}$ and generating a reward $R_t$. The specific implementation process is depicted in Figure \ref{fig:controller_worker_architecture}. Its design adheres to the principles of decoupling and structural plasticity, ensuring that the agent can achieve macro-control over the search process without interfering with the underlying search efficiency.

\section{Detailed Implementation}

\subsection{The Neural Controller}

To address the non-Markovian characteristics of the evolutionary search process—where the optimal action choice depends not only on the current state but also intimately on historical improvement trends and stagnation information—we adopt the Recurrent Proximal Policy Optimization (Recurrent PPO) algorithm, integrated with Long Short-Term Memory (LSTM) networks, as the high-level controller.

For state representation and memory, the controller accepts search states as time-series inputs. Unlike standard feed-forward networks, the LSTM extracts cross-timestep evolutionary features by maintaining a memory of the entire trajectory. Let $o_t$ denote the observation vector (feature state) at generation $t$. The LSTM updates its hidden state $h_t$ and cell state $c_t$ as follows:
\begin{equation} \label{eq:lstm}
    (h_t, c_t) = \text{LSTM}(o_t, h_{t-1}, c_{t-1}; \theta_{enc})
\end{equation}
Here, $h_t$ serves as a compact representation (or belief state) of the entire history $\tau_t = \{o_0, \dots, o_t\}$, effectively converting the non-Markovian observation sequence into a Markovian latent state suitable for subsequent decision-making.

For policy generation and action mapping, the policy network $\pi_\theta$ outputs a continuous action vector based on the history embedding $h_t$. We model the stochastic policy as a multivariate Gaussian distribution:
\begin{equation} \label{eq:policy}
    a_t \sim \pi_\theta(\cdot | h_t) = \mathcal{N}(\mu_\theta(h_t), \Sigma_\theta(h_t))
\end{equation}
where $\mu_\theta$ and $\Sigma_\theta$ are neural networks mapping $h_t$ to the mean and (co)variance of the action space, respectively. The sampled raw action $a_t$ is then mapped to specific GA parameters. Certain action dimensions are discretized for structural parameters (e.g., Operator Modes 0/1/2 and Tournament Size $TS$), while the remaining dimensions are directly mapped to continuous numerical parameters (e.g., Crossover Rate $CR$, Mutation Rate $MR$).

Regarding the optimization objective, this memory-equipped decision-making mechanism enables the controller to perceive long-term dynamic information. To train the controller, we maximize the PPO objective function, which incorporates a clipping mechanism to ensure training stability:
\begin{equation} \label{eq:ppo_loss}
    L^{CLIP}(\theta) = \hat{\mathbb{E}}_t \left[ \min(r_t(\theta)\hat{A}_t, \text{clip}(r_t(\theta), 1-\epsilon, 1+\epsilon)\hat{A}_t) \right]
\end{equation}
where the probability ratio $r_t(\theta)$ is explicitly conditioned on the historical context maintained by the LSTM:
\begin{equation}
    r_t(\theta) = \frac{\pi_\theta(a_t | h_t)}{\pi_{\theta_{old}}(a_t | h_{t}^{old})}
\end{equation}
By optimizing this objective, the controller learns to perform multi-stage global switching and parameter adaptation based on the inferred search dynamics.

\subsection{The High-Performance Evolutionary Worker}
This module constitutes a high-performance Genetic Algorithm, accelerated using Numba JIT, responsible for executing computationally intensive combinatorial optimization tasks. Specifically, it incorporates a dual-level adaptive mechanism: at the numerical level, the algorithm dynamically adjusts operator intensity and selection pressure based on the parameters output by the controller; at the structural level, it integrates dynamic population management and multi-mode operator switching to adapt to the characteristics of various periods in the evolutionary process.

\textbf{Dynamic Population Management} enables the algorithm to flexibly adjust the population size according to the current evolutionary stage and performance. A smaller population is utilized in the early stages to improve computational efficiency and convergence speed, while the population size is gradually expanded in later stages or when improvement slows, to enhance the exploration capability of the search space.

\textbf{Operator Switching} defines three distinct structural modes to handle different evolutionary stages:
\begin{itemize}
    \item \textbf{Mode 0 (Initial Exploration/Stationary Stage):} Applicable to the early search phase. It primarily combines the Inversion operator with one 2-opt operation to rapidly improve path quality while maintaining solution diversity.
    \item \textbf{Mode 1 (Intensified Exploration Stage):} Applicable when the search approaches a local optimum or enters a stagnation phase. It employs the Double-Bridge operator combined with three 2-Opt operations to increase path perturbation capability, thereby facilitating escape from local optima.
    \item \textbf{Mode 2 (Fine-grained Convergence Stage):} Activated in the late convergence phase. It disables random mutation and focuses on five 2-Opt local optimization operations to achieve fine-grained adjustment of the path, further approximating the optimal solution.
\end{itemize}

\begin{figure}[h!]
    \centering
    \includegraphics[width=0.7\linewidth]{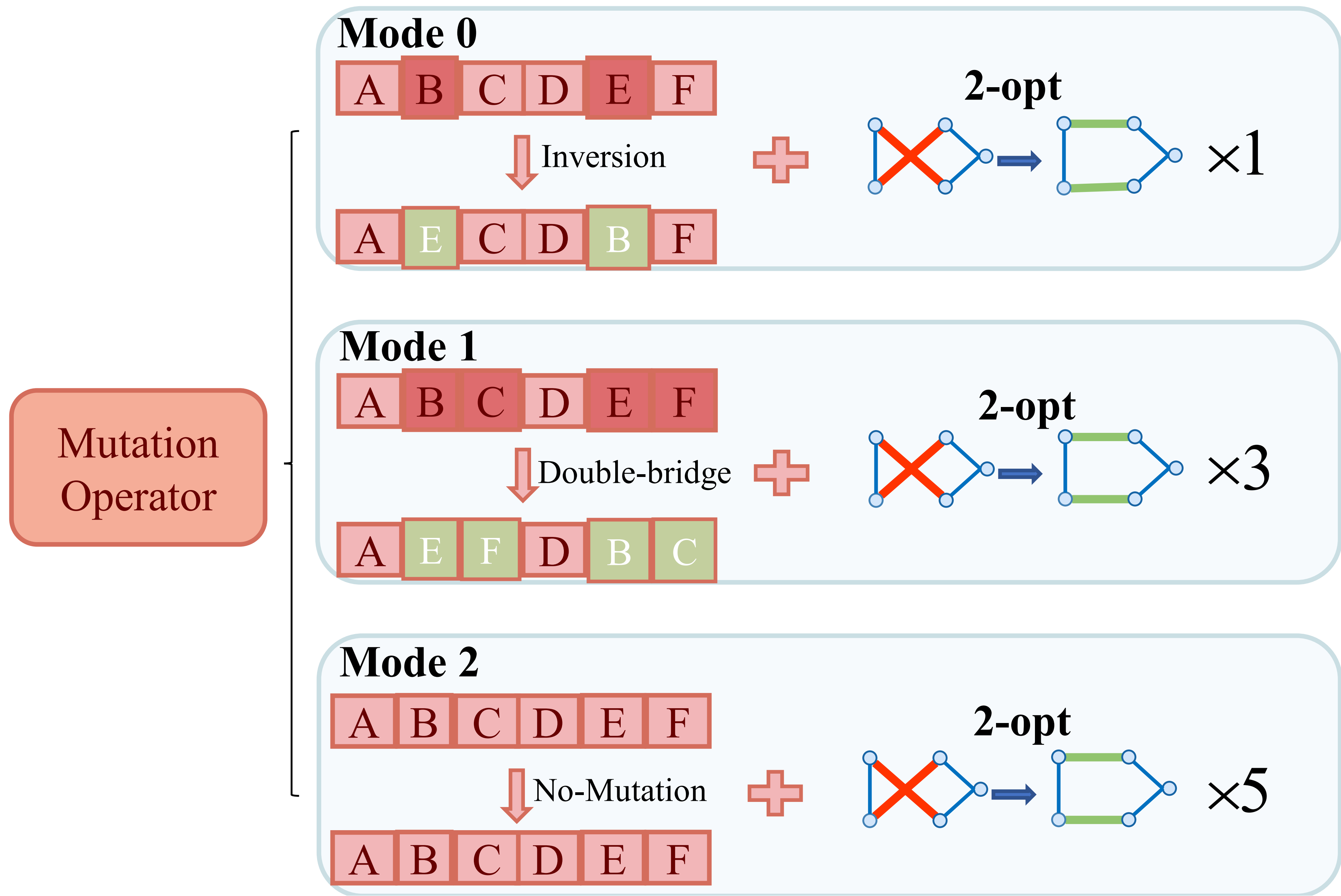} 
    \caption{Mutation Operator}
    \label{fig:mutation_operator} 
\end{figure}

Their specific working principles are illustrated in Figure \ref{fig:mutation_operator}.

\begin{figure}[htbp]
    \centering
    \includegraphics[width=0.7\linewidth]{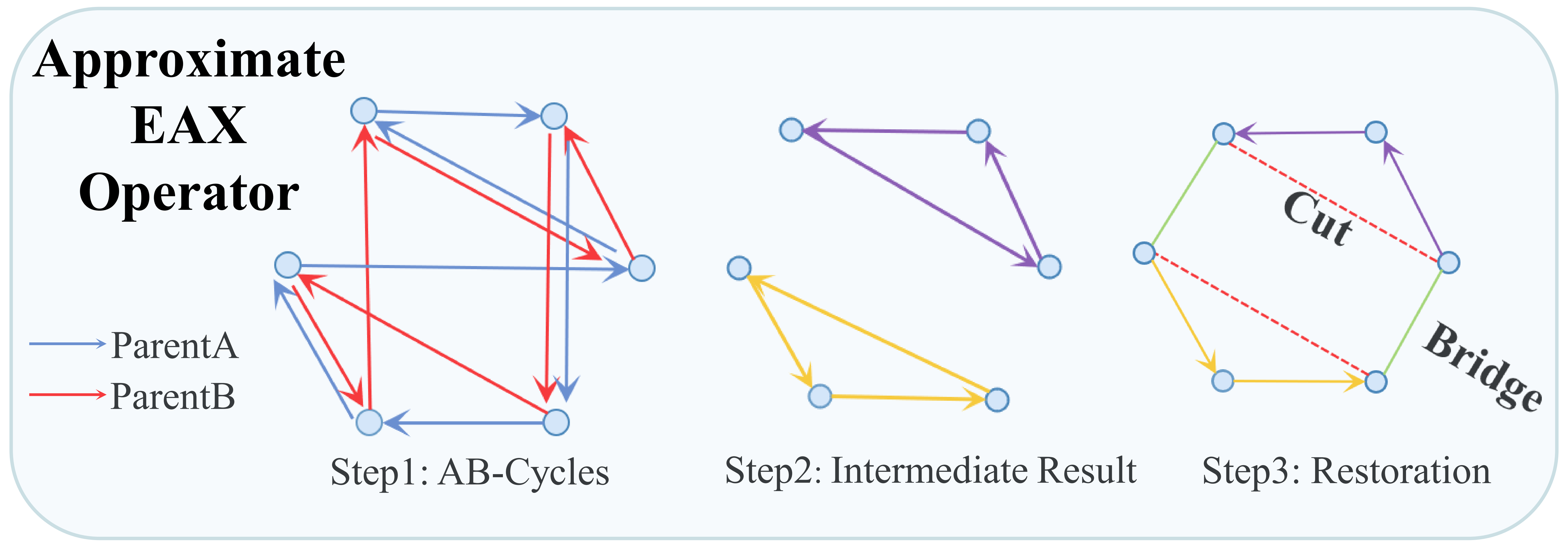}
    \caption{Crossover Operator}
    \label{fig:crossover_operator} 
\end{figure}

For the crossover operator, we employ an approximate EAX operator, whose working principle is illustrated in Figure \ref{fig:crossover_operator}.

Furthermore, when the evolutionary search yields no improvement for a prolonged period (e.g., every hundred generations) and conventional mode switching fails to significantly boost performance, a catastrophic restart mechanism is judiciously introduced to reset the solution space, thereby restoring diversity and search vitality. As a supplementary design to enhance robustness, this catastrophic restart mechanism is implemented independently of the staged operator modes.

\subsection{Training Strategy}
\begin{figure}[t!] 
    \centering
    \includegraphics[width=0.5\linewidth]{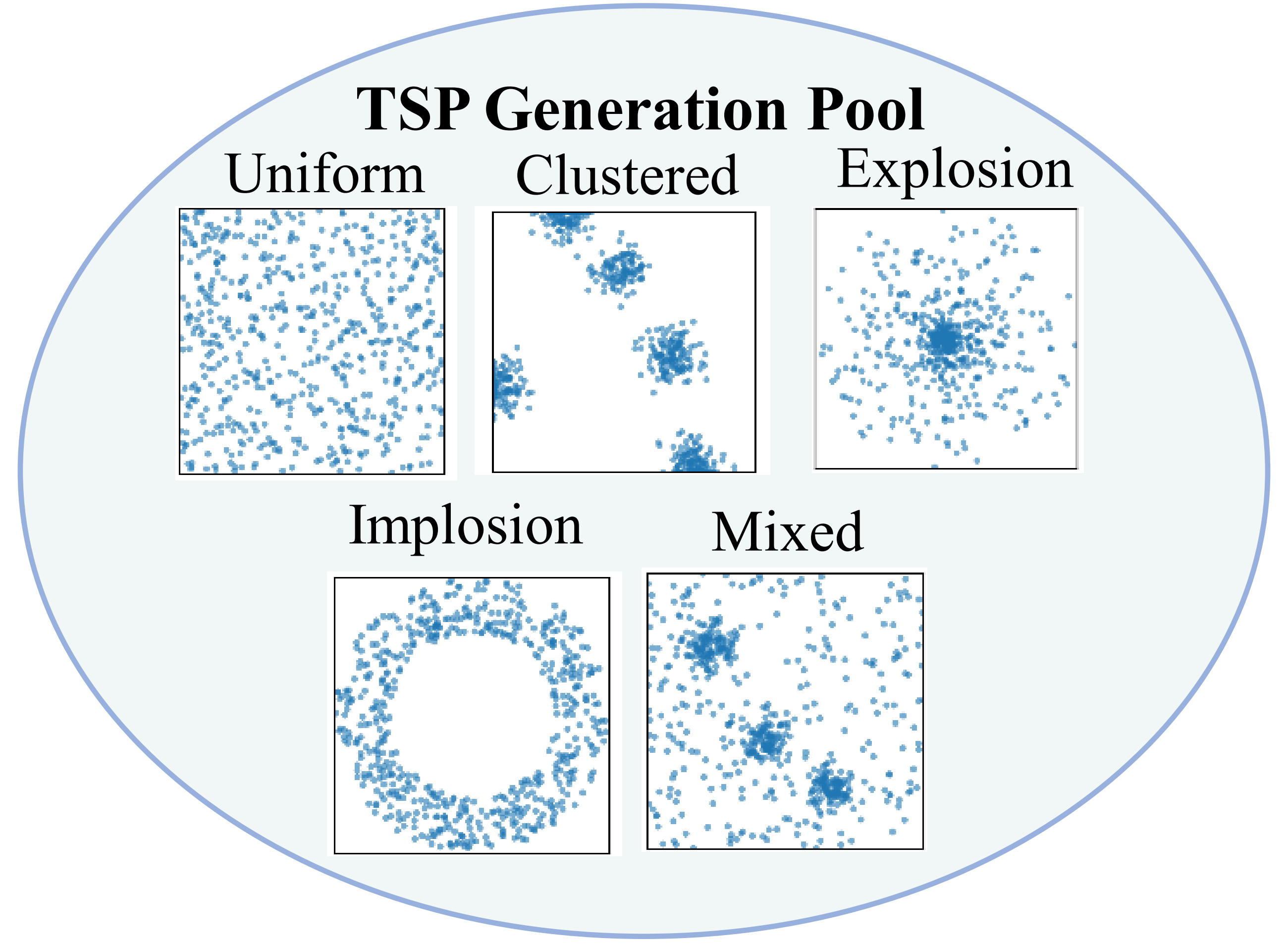}
    \caption{Examples of TSP Instances Generated by the TSP Generator Pool} 
    \label{fig:tsp_generator_pool} 
\end{figure}

To train a control policy with robust generalization capabilities, we adopt an online curriculum learning scheme, which effectively mitigates the cold-start problem in reinforcement learning. The entire training process is divided into four progressively advancing stages: starting from simple instances with 50–150 cities, gradually increasing to medium-scale problems with 400–800 cities, and finally encompassing large-scale problems with up to 1200 cities. To enhance the model's adaptability to diverse topological structures, a TSP Generator Pool was specifically constructed, as illustrated in Figure \ref{fig:tsp_generator_pool}. This pool can generate synthetic examples with various spatial distributions, such as uniform, clustered, explosion, implosion, and mixed types, thereby compelling the controller to learn general switching strategies independent of specific structures.

In terms of engineering implementation, the combination of CurriculumCallback and set\_attr mechanisms ensures the synchronization of curriculum progress in a multi-process environment, thereby guaranteeing the stability and efficiency of distributed training. During the inference phase, the LSTM hidden states are continuously passed between steps to maintain contextual information. The controller then operates in a deterministic manner on test samples, achieving plug-and-play zero-shot generalization capabilities without any online fine-tuning.

\section{Experiments and Analysis}

In this section, our experiments are divided into two main parts: the first compares the performance of our DRLGA with a Static GA, while the second is an ablation study investigating whether numerical or structural parameters have a more significant impact on GA performance.

\textbf{Problem Setting} For performance comparison, we selected large-scale instances: fl1577, u2152, pcb3038, fnl4461, and rl5915 from \textbf{TSPLIB} to test the performance of DRLGA against a static GA. For the ablation study, we selected small-scale (kroA100), medium-scale (lin318, pcb442), and large-scale (rat783, pr1002) instances from \textbf{TSPLIB}. 

\textbf{Training.} The proposed DRLGA model is trained on TSP instances ranging from 50 to 1200 cities, sampled from a custom TSP Generator Pool comprising diverse topologies to ensure training stability. Optimization is performed using Adam with an initial learning rate of $3 \times 10^{-4}$, adhering to a linear decay schedule based on training progress. Leveraging the Recurrent PPO architecture with a batch size of 2048, the model trains across 64 parallel environments for 5 million timesteps. This computationally intensive process was executed on a high-performance node equipped with a 64-core AMD EPYC 7452 CPU (@2.35GHz), DDR4 memory, and InfiniBand high-speed interconnects to maximize parallel efficiency.

\textbf{Baselines} We compared the DRLGA against a highly optimized Static GA baseline, which utilizes the exact same Numba-accelerated kernel but operates with fixed, empirically tuned parameters (CR=0.8, MR=0.2, Population Size=500, Approximate EAX crossover operator + Mode 0, with Mode 2 activated every 50 generations).

\textbf{Metrics} We report the optimality gap, inference time, and dynamic parameter trajectories (for both structural and numerical parameters) in Experiment A to analyze the agent’s decision-making process. Experiment B validates the method’s robustness through average convergence trends and statistical box plots. The optimality gap is calculated as $\dfrac{L_{pred}-L_{opt}}{L_{opt}}\times 100\%$, serving as the standard measure of approximation quality. Each experiment was repeated 30 times independently with different random seeds to mitigate the impact of randomness. We report the mean metrics to evaluate the general performance. For experiment data from all test sets mentioned in this experiment, please refer to \url{https://github.com/StarDream1314/DRLGA-TSP}.

\subsection{Performance Comparison: DRLGA vs. Static GA}
As shown in Table \ref{tab:performance_comparison}, DRLGA demonstrates superior asymptotic precision across all instances. We acknowledge that DRLGA incurs a higher computational overhead, with inference times averaging approximately two times that of the Static GA. However, it is crucial to emphasize that the primary objective of this study is not to engineer a runtime-optimized commercial solver, but rather to validate the feasibility of the DRL-driven hyper-heuristic framework and to investigate the distinct roles of structural versus numerical parameters. The significant reduction in the optimality gap serves as a proof of concept, demonstrating that the learned adaptive policies can unlock performance gains that static configurations cannot achieve, prioritizing scientific insight into evolutionary dynamics over raw computational speed.

\begin{table}[htbp]
    \centering
    \caption{Performance Comparison}
    \label{tab:performance_comparison} 
    
    \renewcommand{\arraystretch}{1.2}
    \setlength{\tabcolsep}{5pt}
    
    \begin{tabular}{l c c c c}
        \toprule
        \multirow{2}{*}{\textbf{Instance}} & \multicolumn{2}{c}{\textbf{Optimality Gap (\%)}} & \multicolumn{2}{c}{\textbf{Inference Time (s)}} \\
        \cmidrule(lr){2-3} \cmidrule(lr){4-5}
         & \textit{Static GA} & \textbf{\textit{DRLGA (Ours)}} & \textit{Static GA} & \textit{DRLGA (Ours)} \\
        \midrule
        
        \textbf{fl1577}  & 9.10  & \textbf{4.44}  & $\approx$183   & $\approx$347 \\ 
        
        \textbf{u2152}   & 13.50 & \textbf{4.57}  & $\approx$426   & $\approx$869 \\
        
        \textbf{pcb3038} & 11.53 & \textbf{4.89}  & $\approx$746   & $\approx$1606 \\
        
        \textbf{fnl4461} & 9.84  & \textbf{6.60}  & $\approx$1,807 & $\approx$3903 \\
        
        \textbf{rl5915}  & 19.19 & \textbf{10.48} & $\approx$2,821 & $\approx$6,240 \\
        
        \bottomrule
    \end{tabular}

\end{table}

\begin{figure}[h]
    \centering
    \includegraphics[width=1\linewidth, keepaspectratio]{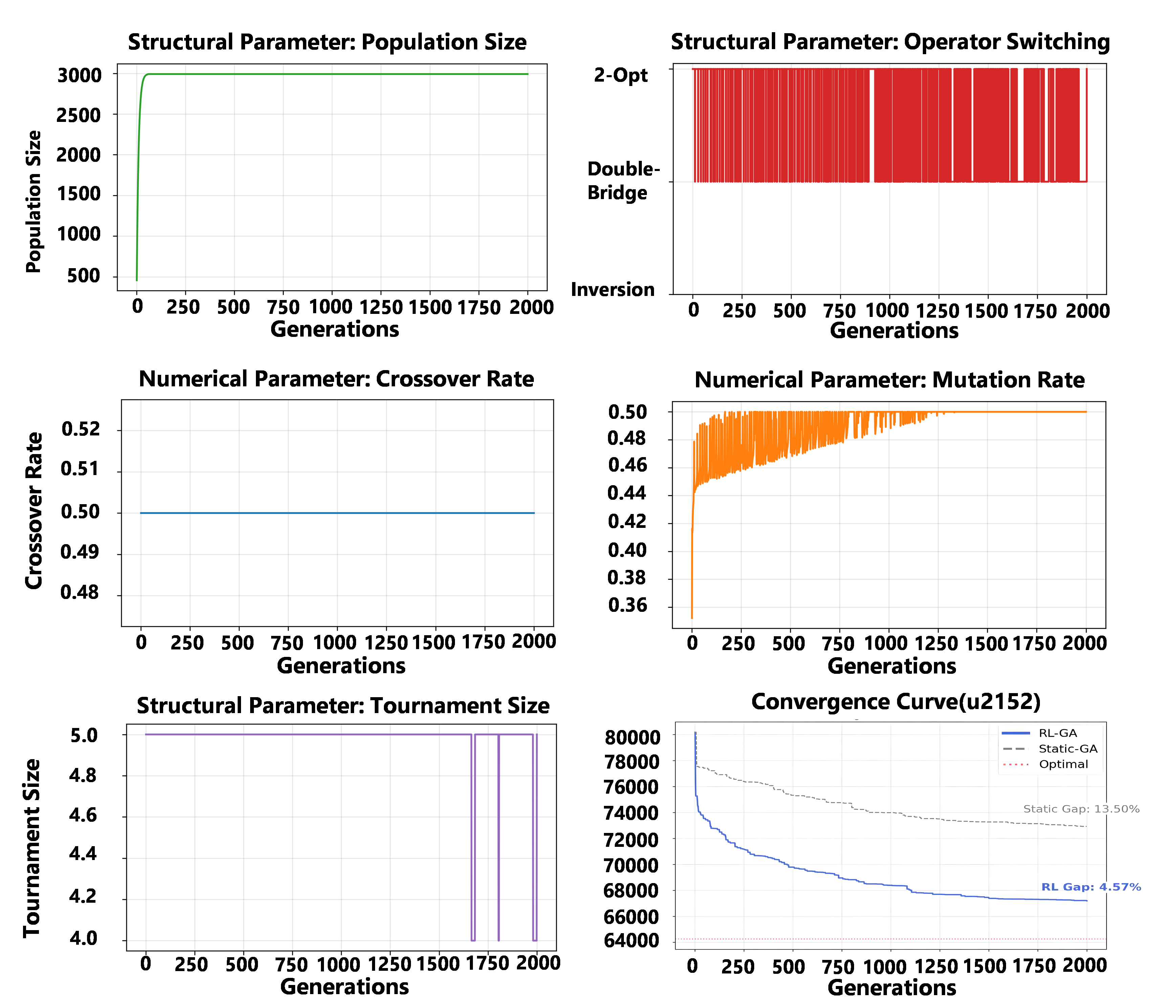}
    \caption{Performance Comparison}
    \label{Performance Comparison} 
\end{figure}

As presented in Figure \ref{Performance Comparison}, the static GA exhibits typical early stagnation (an "L-shaped" convergence curve), failing to escape local optima. In contrast, DRLGA maintains a sustained "step-wise" descent in its optimality gap. Notably, on u2152, DRLGA reduces the optimality gap from 13.50\% (Static GA) to 4.57\%; for the largest instance, rl5915, the gap drops from 19.19\% to 10.48\%. This confirms the agent’s superior ability to navigate complex fitness landscapes where static parameters prove inadequate.

The parameter trajectories reveal three intelligent mechanisms driving this superior performance:

1. \textbf{Resource Saturation}: The agent maximizes population size early in the search process to ensure sufficient information-carrying capacity for large-scale problem exploration.

2. \textbf{Hyper-heuristic Oscillation}: The agent frequently switches between operators (Mode 1 and Mode 2), leveraging topological complementarity to disrupt population homogeneity and prevent premature convergence.

3. \textbf{Inverse Annealing}: This refers to a counter-intuitive behavior where mutation rates increase towards the end of evolution (approaching 0.5). We interpret this as an adaptive response to the late-stage “diversity crisis,” forcefully injecting entropy to break stagnation in large-scale instances. 

It is noteworthy that DRLGA achieves robust zero-shot generalization, embodying the principle of "Train Small, Solve Large." Trained solely on synthetic datasets with $N \in [50, 1200]$, the agent performs zero-shot inference on TSPLIB instances with up to $N=5915$ cities. Despite a five-fold increase in problem size, the DRLGA agent avoids the "Out-of-Distribution" failure common in constructive neural heuristics. On rl5915, it achieves an approximately $50\%$ improvement over the baseline, demonstrating that the learned control logic is robust against variations in problem dimensionality. By observing normalized evolutionary statistics rather than raw node coordinates, the state representation becomes scale-invariant. Consequently, the learned meta-control policies (e.g., "Inverse Annealing" during stagnation) function as universal optimization laws, applicable regardless of the problem size.

\subsection{Ablation Study: Numerical Parameters vs. Structural Parameters}
\begin{figure}[htbp!]
    \centering
    
 \begin{subfigure}[b]{\linewidth}
    \centering
    \includegraphics[width=0.9\linewidth]{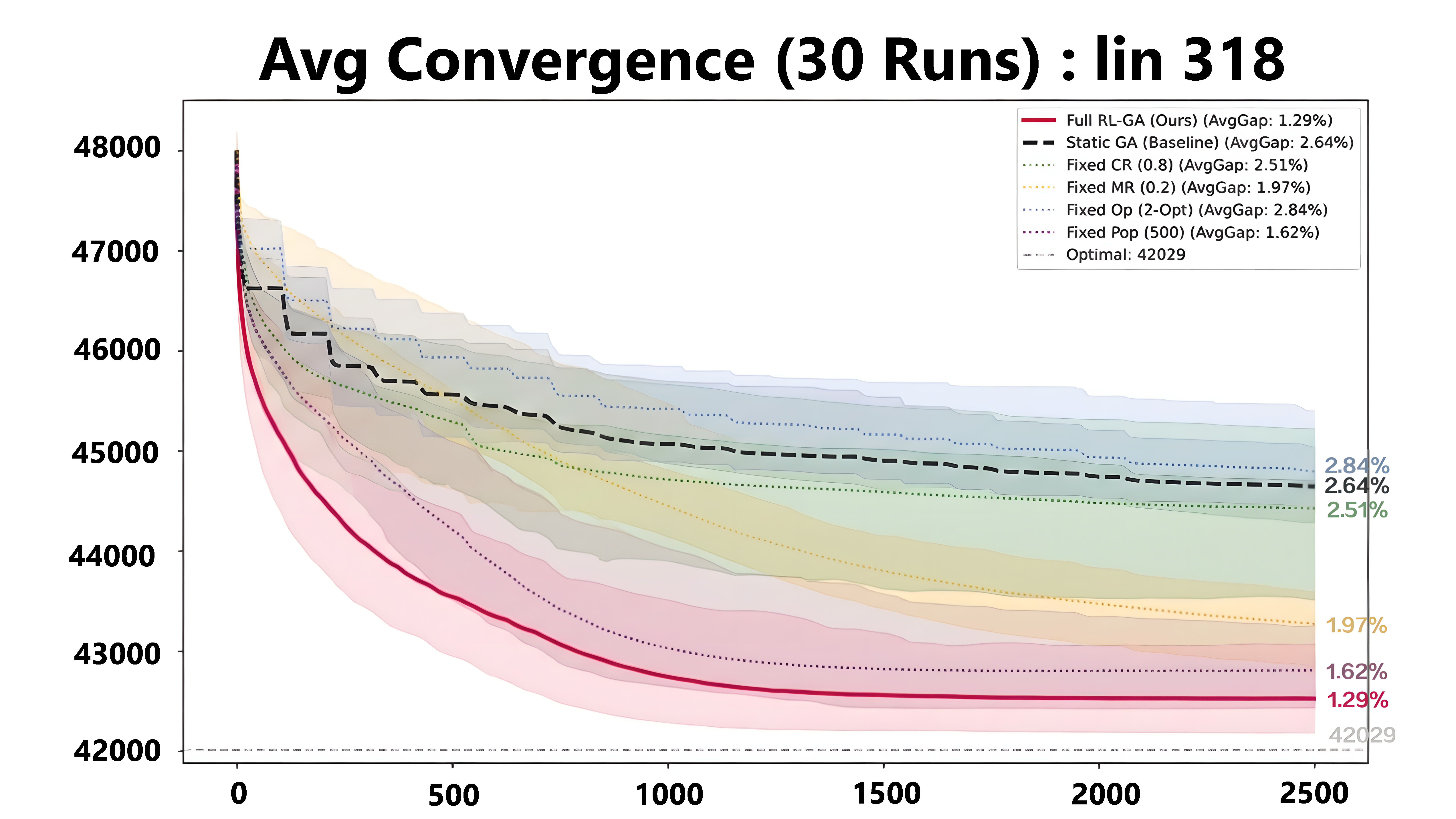}
    \caption{Convergence Curve}
    \label{fig:curve_lin318}
\end{subfigure}

    \vspace{0.5cm} 
    
    \begin{subfigure}[b]{1\linewidth}
        \centering
        \includegraphics[width=0.9\linewidth, keepaspectratio]{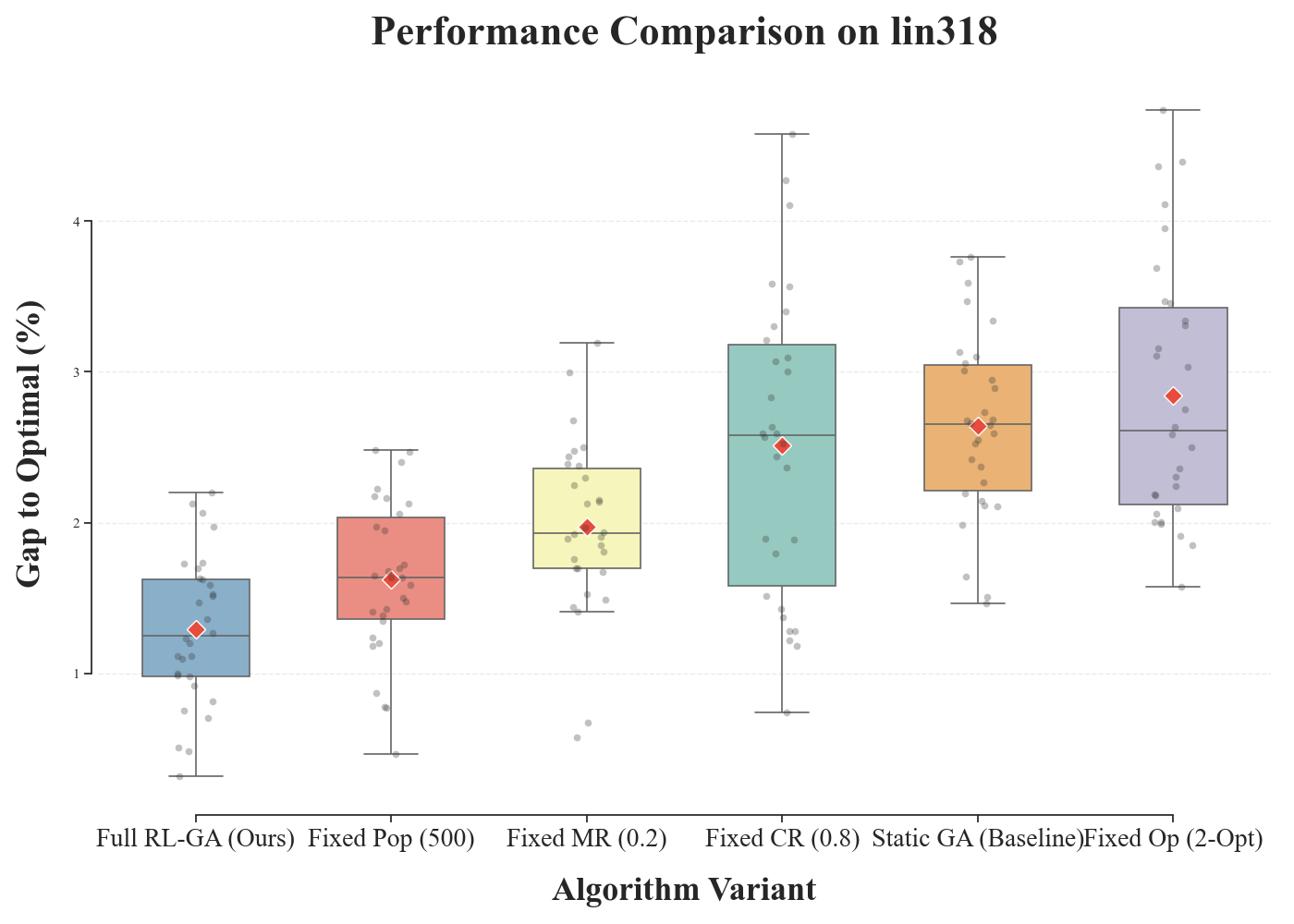}
        \caption{Box Plot}
        \label{fig:box_lin318}
    \end{subfigure}

    \caption{Analysis of lin318 Instance}
    \label{fig:lin318_ablation}
\end{figure}

To disentangle the individual impacts of distinct control variables, we performed a systematic ablation study by classifying hyperparameters into structural (population size, operator mode) and numerical (crossover, mutation rates) categories. The results are partly shown in Figure \ref{fig:lin318_ablation}.

\begin{table}[htbp!]
    \centering
    \caption{Ablation Study}
    \label{tab:ablation_study}
    
    \renewcommand{\arraystretch}{1.25}
    
    \resizebox{\columnwidth}{!}{
        \begin{tabular}{llcccc}
            \toprule
            \multirow{2}{*}{\textbf{Dataset}} & \multirow{2}{*}{\textbf{Experiment}} & \multicolumn{4}{c}{\textbf{Gap (\%)}} \\
            \cmidrule(lr){3-6}
             & & \textbf{Mean} & \textbf{Std} & \textbf{Min} & \textbf{Max} \\
            \midrule
    
            \multirow{6}{*}{\textbf{kroA100}} 
              & Fixed CR (0.8)       & 0.07 & 0.02 & 0.06 & 0.12 \\
              & Fixed MR (0.2)       & 0.08 & 0.03 & 0.06 & 0.12 \\
              & Fixed Op (2-Opt)     & 0.18 & 0.06 & 0.12 & 0.28 \\
              & Fixed Pop (500)      & 0.08 & 0.03 & 0.06 & 0.12 \\
              & \textbf{Full DRLGA (Ours)} & \textbf{0.08} & \textbf{0.03} & \textbf{0.06} & \textbf{0.12} \\ 
              & Static GA (Baseline) & 0.18 & 0.06 & 0.12 & 0.28 \\
            \midrule
            
            \multirow{6}{*}{\textbf{lin318}} 
              & Fixed CR (0.8)       & 2.39 & 0.96 & 0.74 & 4.58 \\
              & Fixed MR (0.2)       & 1.93 & 0.69 & 0.58 & 3.19 \\
              & Fixed Op (2-Opt)     & 2.90 & 0.79 & 1.57 & 4.73 \\
              & Fixed Pop (500)      & 1.71 & 0.54 & 0.47 & 2.48 \\
              & \textbf{Full DRLGA (Ours)} & \textbf{1.35} & \textbf{0.50} & \textbf{0.32} & \textbf{2.20} \\ 
              & Static GA (Baseline) & 2.58 & 0.63 & 1.46 & 3.76 \\
            \midrule
    
            \multirow{6}{*}{\textbf{pcb442}} 
              & Fixed CR (0.8)       & 1.19 & 0.47 & 0.47 & 2.80 \\
              & Fixed MR (0.2)       & 1.29 & 0.52 & 0.52 & 2.31 \\
              & Fixed Op (2-Opt)     & 3.69 & 0.78 & 2.04 & 5.23 \\
              & Fixed Pop (500)      & 1.19 & 0.50 & 0.34 & 2.35 \\
              & \textbf{Full DRLGA (Ours)} & \textbf{1.02} & \textbf{0.44} & \textbf{0.35} & \textbf{2.28} \\ 
              & Static GA (Baseline) & 2.65 & 0.82 & 0.77 & 4.51 \\
            \midrule
    
            \multirow{6}{*}{\textbf{pr1002}} 
              & Fixed CR (0.8)       & 3.80 & 0.79 & 2.47 & 6.03 \\
              & Fixed MR (0.2)       & 5.04 & 0.65 & 3.32 & 6.40 \\
              & Fixed Op (2-Opt)     & 6.36 & 0.88 & 4.88 & 8.40 \\
              & Fixed Pop (500)      & 4.38 & 0.72 & 2.83 & 6.00 \\
              & \textbf{Full DRLGA (Ours)} & \textbf{3.85} & \textbf{0.80} & \textbf{2.68} & \textbf{5.90} \\ 
              & Static GA (Baseline) & 5.16 & 0.77 & 3.66 & 6.82 \\
            \bottomrule
        \end{tabular}
    }
\end{table}

As detailed in Table \ref{tab:ablation_study}, a distinct hierarchy emerges from the empirical data: structural adaptability functions as the dominant factor for scalability, whereas numerical parameters occupy a secondary role primarily focused on fine-tuning. Performance degradation was most acute when the algorithm was confined to a static operator ("Fixed Op" setting), particularly on large-scale instances. This evidence confirms that dependence on a singular neighborhood topology (e.g., 2-Opt) invariably results in stagnation at local optima. A similar trend was observed with fixed population sizes ("Fixed Pop" setting), highlighting a deficiency in information-carrying capacity. This validates the necessity of dynamic population scaling to sustain genetic diversity and exploration potential in high-dimensional search spaces.

Conversely, constraints on numerical parameters produced a more moderate impact. While the "Fixed CR" variant displayed sensitivity on medium-sized problems like lin318 (2.39\% gap), the "Fixed MR" variant showed unexpected robustness on the massive pr1002 instance. By maintaining a gap of 5.04\%—significantly outperforming the 6.36\% gap of the structural failure ("Fixed Op" setting)—the results suggest a critical distinction: while mutation intensity regulates convergence velocity, it is less prone to inducing irreversible stagnation than structural rigidity. The search process may proceed more slowly with static mutation, but it can be effectively halted by an inflexible topology. Ultimately, the Full DRLGA secured the lowest mean gap across all datasets (e.g., 1.02\% on pcb442). This superiority stems from a synergistic orchestration: the agent utilizes structural shifts to break out of basins of attraction, while reserving numerical adjustments to accelerate exploitation within promising regions.

\section{Conclusion and Future Work}

This paper proposed a dual-level adaptive Genetic Algorithm framework, reinforced by Recurrent PPO, which decouples the control of numerical and structural parameters for fine-grained management of the evolutionary search process. Through systematic experimentation and analysis, we have not only validated the efficiency of this framework in solving large-scale Traveling Salesman Problems but also derived a critical insight that challenges traditional perspectives in evolutionary computation: when scaling to complex, high-dimensional optimization tasks, adaptive structural management (such as dynamic population size and operator switching) has a far more decisive impact on performance than traditional numerical probability fine-tuning. This work establishes a new paradigm for designing scalable, intelligent optimization systems by elevating adaptation from solely the numerical level to encompass the structural level. Future work will extend this framework to constrained domains like the Vehicle Routing Problem (VRP) to test the transferability of learned meta-policies. Additionally, we aim to integrate automatically generated operators and apply this control mechanism to other evolutionary algorithms, such as Differential Evolution, to further generalize the findings on structural plasticity.


\bibliographystyle{IEEEtran}
\bibliography{references}
\end{document}